\documentclass[11pt]{article}

\usepackage[utf8]{inputenc}
\usepackage{arxiv}

\usepackage{cite}
\usepackage{amsmath,amssymb,amsfonts}
\usepackage{algorithmic}
\usepackage{graphicx}
\usepackage{textcomp}
\usepackage{xcolor}
\usepackage{multirow}
\usepackage{booktabs}
\usepackage{array}
\usepackage{float}

\begin{document}

\title{Multidimensional Analysis of Specific Language Impairment Using Unsupervised Learning Through PCA and Clustering}

\author{Niruthiha Selvanayagam\\
College of Engineering\\
Northeastern University\\
Boston, MA, USA\\
\texttt{selvanayagam.n@northeastern.edu}}

\maketitle

\begin{abstract}
Specific Language Impairment (SLI) affects approximately 7 percent of children, presenting as isolated language deficits despite normal cognitive abilities, sensory systems, and supportive environments. Traditional diagnostic approaches often rely on standardized assessments, which may overlook subtle developmental patterns. This study aims to identify natural language development trajectories in children with and without SLI using unsupervised machine learning techniques, providing insights for early identification and targeted interventions. Narrative samples from 1,163 children aged 4-16 years across three corpora (Conti-Ramsden 4, ENNI, and Gillam) were analyzed using Principal Component Analysis (PCA) and clustering. A total of 64 linguistic features were evaluated to uncover developmental trajectories and distinguish linguistic profiles. Two primary clusters emerged: (1) high language production with low SLI prevalence, and (2) limited production but higher syntactic complexity with higher SLI prevalence. Additionally, boundary cases exhibited intermediate traits, supporting a continuum model of language abilities. Findings suggest SLI manifests primarily through reduced production capacity rather than syntactic complexity deficits. The results challenge categorical diagnostic frameworks and highlight the potential of unsupervised learning techniques for refining diagnostic criteria and intervention strategies.
\end{abstract}


\section{Introduction}

From a child's first words to complex conversations, language acquisition marks a remarkable journey in human development. Most children master intricate linguistic patterns naturally, yet a significant number face an intriguing developmental challenge: they struggle to acquire language skills despite normal cognitive abilities, intact sensory systems, and supportive environments. This condition, known as Specific Language Impairment (SLI), affects approximately 7\% of children\cite{tomblin1997}, making it more common than autism spectrum disorders or dyslexia, yet it remains poorly understood.

The study of SLI has evolved significantly over the past decades, shifting from simple descriptive approaches to sophisticated analyses of language development patterns. Studies have shown SLI manifests in key linguistic domains, with research highlighting impacts on the complexity of language and quality of utterances, particularly in spontaneous narrative contexts\cite{arena2024sli}. However, these studies typically rely on predetermined developmental milestones and standardized assessments, potentially overlooking subtle patterns that could revolutionize our understanding of language development.

Traditional diagnostic approaches for SLI often rely on categorical classifications based on standardized assessments. These methods, while practical for clinical settings, may miss the nuanced developmental trajectories that characterize language acquisition. This study aims to identify natural language development patterns in children with and without SLI through unsupervised machine learning techniques, potentially revealing insights for early identification and targeted interventions.

By analyzing narrative samples from 1,163 children aged 4-16 years across three well-established corpora and evaluating 64 distinct linguistic features, this research applies Principal Component Analysis (PCA) and clustering techniques to uncover developmental trajectories and distinguish linguistic profiles. This approach moves beyond traditional assessment methods to explore whether SLI manifests as part of a continuous spectrum of language abilities rather than a discrete diagnostic category.

A particularly compelling aspect of SLI research lies in its unique manifestation: unlike language delays associated with hearing impairment, cognitive deficits, or neurological conditions, SLI presents as an isolated impairment in language development. This specificity makes it an ideal candidate for studying the fundamental mechanisms of language acquisition. Yet, current diagnostic frameworks struggle to capture its heterogeneous nature, leading to delayed identification and intervention. This study aims to address this challenge by providing a data-driven framework for understanding the multidimensional nature of language impairment.

\section{Related Work}

Specific Language Impairment (SLI) affects approximately 7\% of children, presenting as isolated language deficits despite normal cognitive abilities and intact sensory systems \cite{leonard2014, tomblin1997}. Traditional clinical approaches have relied on standardized assessments evaluating morphological development, grammatical accuracy, and syntactic complexity \cite{rice2020}. However, these methods often fail to capture the interconnected nature of language development, potentially missing subtle patterns crucial for early intervention \cite{bishop2016}.

Recent work by Huang et al. (2022) demonstrated the effectiveness of supervised learning approaches for SLI classification\cite{huang2022machine}. Their comparative analysis of Logistic Regression, AdaBoost, Random Forest, and BPNN models revealed BPNN as the most effective classifier, achieving consistent performance in SLI detection. While their work established the viability of machine learning for clinical applications, their supervised approach necessarily relied on predetermined diagnostic categories. This research extends beyond classification to explore natural language development patterns through unsupervised learning, potentially revealing developmental trajectories that binary classification might miss.

The field has evolved from manual analysis to computational methods, supervised learning techniques have successfully identified specific linguistic markers \cite{gabani2011}, but their reliance on predefined categories limits their ability to discover novel developmental patterns. Unsupervised learning offers a promising alternative, with recent work demonstrating its potential in identifying natural language patterns in typical development \cite{roy2019}.

While machine learning has shown promise in clinical linguistics \cite{duran2018}, most applications focus on classification rather than pattern discovery. Dimensionality reduction techniques, though well-established in linguistics, have been primarily limited to feature selection and visualization in developmental language disorders \cite{wetherell2007}.

Current research gaps include:
\begin{itemize}
    \item Limited application of unsupervised learning to SLI analysis
    \item Lack of methods for identifying natural developmental patterns
    \item Insufficient integration of multiple linguistic domains
\end{itemize}

This work addresses these limitations through a novel computational framework combining PCA with clustering techniques, enabling the discovery of natural developmental trajectories while integrating multiple linguistic features. This approach moves beyond traditional categorical assessments to uncover underlying patterns in language development, potentially improving early identification and intervention strategies.

\section{Dataset Description}

The data in this study comprises narrative language samples from three well-established corpora in the CHILDES (Child Language Data Exchange System) database \cite{okeeffe2020}. The combined dataset consists of 1,163 narrative samples from children with and without Specific Language Impairment (SLI), representing diverse age groups and geographical locations.

\subsection{Corpus Composition}
The Conti-Ramsden corpus \cite{ContiRamsden2024} includes longitudinal data from three children with SLI and their younger siblings, providing detailed mother-child interaction analyses \cite{contiramsden1991}. The Conti-Ramsden 4 corpus includes 118 British adolescents (99 typically developing, 19 with SLI) aged 13.10 to 15.90 years. These samples were collected using Mayer's wordless picture book "Frog, Where Are You?" as a narrative prompt. Participants viewed the picture book independently before retelling the story, with instructions to use past tense. When participants deviated to present tense, a maximum of two prompts ("What happened next?") were provided to encourage past tense usage.

The Edmonton Narrative Norms Instrument (ENNI) corpus comprises 377 samples from Canadian children aged 4 to 9 years (300 typically developing, 77 with SLI) \cite{ENNI2024} . The ENNI protocol employed two wordless picture stories of increasing complexity, with examiners controlling page turns to ensure systematic progression through the narrative. Participants received practice opportunities with a training story, during which explicit prompts were provided to establish task expectations \cite{schneider2006}.

The Gillam corpus, based on the Test of Narrative Language (TNL), is the largest subset with 770 samples from U.S. children aged 5 to 12 years (520 typically developing, 250 with SLI) \cite{gillam2004test}. The TNL protocol incorporated four distinct storytelling tasks: one script-based story recall and three wordless picture narratives. These tasks were designed with progressive complexity, beginning with sequential picture stories and advancing to single-picture prompts requiring greater inferential and narrative skills.

\subsection{Demographic Distribution}
Age distribution across the combined dataset shows significant coverage of critical developmental periods, with particular concentration in the 5-12 year range where language impairments often become most apparent. The sample includes balanced gender representation, though 119 cases (10.2\%) have missing gender information. Geographic distribution across three English-speaking countries (UK, Canada, USA) enhances the dataset's generalizability while maintaining linguistic consistency through English-only samples.

\subsection{Clinical Classification}
The SLI diagnosis in all three corpora followed standard clinical criteria: language difficulties in the absence of hearing impairment, neurological dysfunction, or intellectual disability. The combined dataset includes 267 children with SLI (23\%) and 896 typically developing children (77\%), reflecting typical clinical prevalence rates. Diagnostic criteria were consistently applied across all three corpora, with language performance at least 1.25 standard deviations below age expectations for SLI classification.

\subsection{Task Characteristics}
Each corpus employed specific elicitation protocols designed to assess narrative abilities:
The Conti-Ramsden 4 protocol emphasized temporal sequencing and past tense usage through its single extended narrative task. The ENNI protocol focused on story grammar and causal relationships through contrasting narrative complexities. The TNL protocol provided the most comprehensive assessment through its multiple task types, enabling evaluation of both recall and generative narrative abilities.

\subsection{Data Validation}
Inter-rater reliability was established for transcription and coding in each corpus (Cohen's $\kappa > 0.80$). All transcripts followed CHILDES conventions, ensuring consistency in morphological coding and utterance segmentation. The combined dataset underwent additional validation to ensure compatibility across corpora, including standardization of transcription formats and verification of clinical classifications.

\subsection{Dataset Features}

This analysis incorporates 64 distinct linguistic features, organized into five primary categories, as detailed in Tables I through V. Table I presents syntactic and error analysis features, including complexity measures and developmental scores. Table II describes morphological development markers, encompassing various grammatical forms and structures. Table III covers language model perplexity measures and utterance structure features, including fluency indicators. Table IV details standardized comparison metrics through Z-score features across different linguistic dimensions.   Finally, Table V contains basic production measures and lexical diversity features, capturing fundamental aspects such as word count and lexical complexity. 

These features were selected to provide comprehensive coverage of language development patterns, ranging from basic production metrics to sophisticated syntactic analysis. Each category systematically captures different aspects of linguistic development, enabling detailed analysis of language impairment patterns.

\section{Feature Processing Methodology}

\subsection{Data Preprocessing}
Initial data cleaning involved addressing missing values, particularly in demographic variables (119 missing entries in the sex variable). Feature standardization was performed to ensure comparability across different scales, with each feature normalized to zero mean and unit variance. This standardization was essential given the diverse nature of the linguistic measures, ranging from simple counts to complex ratios and scores.

\subsection{Feature Selection and Transformation}

The feature set was refined through correlation analysis to address multicollinearity, particularly among related linguistic measures. Features were retained based on both their statistical independence and theoretical significance in language development research. This process resulted in a final set of 59 features for the main analysis.

The resultant dataset provides a comprehensive basis for analyzing language development patterns, with special attention to the distinctions between typically developing children and those with Specific Language Impairment. The diverse age ranges and geographical distribution of the samples enhance the generalizability of the findings, while the detailed feature set enables fine-grained analysis of language development patterns.

\section{Analysis Methods}

\subsection{Dimensionality Reduction}

Principal Component Analysis (PCA) was chosen as the primary dimensionality reduction technique after careful evaluation of several alternatives. While factor analysis offered potential benefits in modeling latent variables, PCA demonstrated superior characteristics for the specific application in analyzing linguistic features. PCA functions by transforming high-dimensional data into a lower-dimensional space while preserving the maximum amount of variance\cite{jaadi2022}. This transformation is achieved by computing eigenvectors of the data's covariance matrix, which become the principal components. While the method can compute as many principal components as there are original variables, dimensionality reduction is accomplished by retaining only the components that explain the majority of variance in the data. In my analysis, this approach proved particularly valuable for handling the complex, interconnected nature of linguistic features while maintaining the interpretability crucial for clinical applications. 

The key advantages of PCA include:
\begin{itemize}
    \item Effective preservation of the underlying variance structure in the data
    \item Superior computational efficiency when handling high-dimensional linguistic features
    \item Robust handling of multicollinearity among language measures
    \item Generation of more clinically interpretable results
\end{itemize}

2) Other Alternatives Considered:
\begin{itemize}
   \item t-SNE was evaluated but rejected due to its non-linear nature making clinical interpretation more challenging
   \item UMAP was tested but showed less stable results across different parameter settings
   \item Linear Discriminant Analysis was inappropriate due to its supervised nature
\end{itemize}

The analysis employed Principal Component Analysis (PCA) for dimensionality reduction, applied to 59 standardized features. The optimal number of components was determined through multiple criteria, including the Kaiser criterion (eigenvalue $> 1$) and explained variance thresholds. The first three principal components explained 48.46\% of the total variance, with the first component accounting for 28.35\%, the second for 13.23\%, and the third for 6.87\%.

The clustering methodology employed a multi-algorithm approach to ensure robust pattern identification. K-means clustering was applied to the principal components, with additional validation through hierarchical clustering and DBSCAN. Cluster stability was quantitatively assessed using silhouette analysis (scores ranging from 0.416 to 0.460) and Adjusted Rand Index (exceeding 0.86 for primary components). A detailed analysis of 59 boundary cases provided insights into gradient patterns in language development.

\subsection{Validation and Implementation}
Statistical validation combined quantitative metrics with clinical assessment of cluster characteristics. The analysis pipeline was implemented in Python, utilizing scikit-learn for machine learning operations, scipy for statistical analyses, and pandas for data manipulation. 

This methodology integrates computational rigor with clinical relevance, enabling objective analysis of language development patterns while maintaining connection to established clinical understanding. The approach allows natural patterns to emerge without imposing predetermined developmental stages, while ensuring reliability through multiple validation methods.

\begin{figure}[htbp]
    \centering
    \includegraphics[width=\linewidth]{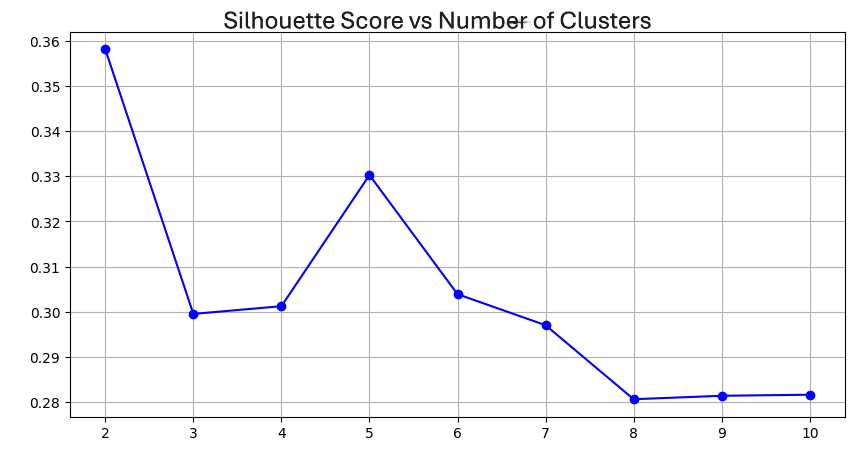}
\caption{Silhouette score analysis showing optimal cluster number determination. Silhouette scores decrease from k=2 (0.36) to k=10 (0.28), with a local maximum at k=5 (0.33). The dominant peak at k=2 indicates optimal cluster separation, while subsequent scores demonstrate diminishing cohesion with increased cluster numbers.
}
\label{fig:enter-label}
\end{figure}

\section{Results}

\subsection{Principal Component Analysis Results}
Principal Component Analysis (PCA) revealed fourteen significant components with eigenvalues exceeding the Kaiser criterion ($\lambda > 1$), which collectively accounted for 83.55\% of the total variance (Table VIII). The optimal number of principal components was determined using multiple criteria:

\begin{itemize}
    \item \textbf{Kaiser Criterion:} Components with an eigenvalue $\lambda > 1$ were retained, resulting in 14 components.
    \item \textbf{Explained Variance Thresholds:}
    \begin{itemize}
        \item 90\% variance explained: 19 components
        \item 95\% variance explained: 26 components
        \item 99\% variance explained: 35 components
    \end{itemize}
    \item \textbf{Elbow Method:} Suggested retaining 2 components.
\end{itemize}

The first three principal components explained a cumulative 48.46\% of the total variance:

\begin{itemize}
    \item \textbf{PC1 (28.35\% of variance):} Showed the highest loadings on morphological words (0.240), total word count (0.239), and syllable count (0.238).
    \item \textbf{PC2 (13.23\% of variance):} Loaded on mean length of utterance in morphemes (0.319), standardized MLU scores (0.315), and verb usage patterns (0.293).
    \item \textbf{PC3 (6.87\% of variance):} Reflected error patterns through word errors (0.284) and perplexity scores (0.270).
\end{itemize}

\begin{figure}[htbp]
    \centering
    \includegraphics[width=1\linewidth]{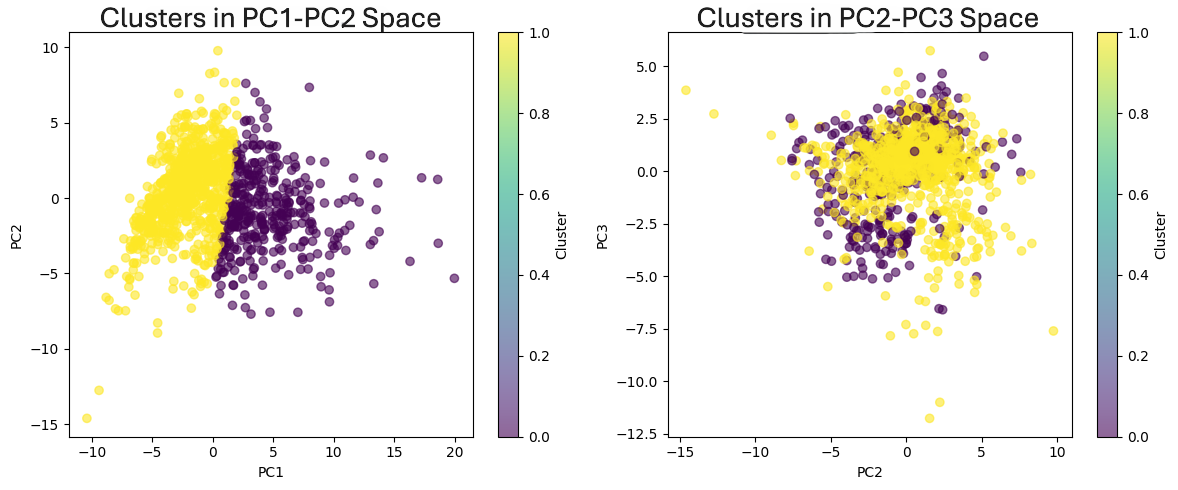}
    \caption{Cluster Visualization in Principal Component Space. The plots show the distribution of two identified clusters (shown in purple and yellow) across different principal component planes. Left: PC1 vs PC2 plot reveals a clear separation between clusters, with Cluster 0 (purple) predominantly in the positive PC1 region and Cluster 1 (yellow) in the negative PC1 region, indicating distinct language production patterns. Right: PC2 vs PC3 plot shows more overlap between clusters, suggesting that the second and third principal components capture variation that is less cluster-specific. These visualizations confirm that PC1 is the primary discriminator between the two clusters, while PC2 and PC3 capture within-cluster variability.}
    \label{fig:enter-label}
\end{figure}

\vspace{1cm}

\begin{table*}[!t]
\caption{Syntactic and Error Analysis Features}
\begin{center}
\begin{tabular}{|p{2.5cm}|p{2cm}|p{11cm}|}
\hline
\textbf{Category} & \textbf{Feature} & \textbf{Description} \\
\hline
\multirow{17}{*}{\parbox{2.5cm}{Syntactic and Error Analysis}} 
& word\_errors & Number of word-level errors \\
& f\_k & Flesch-Kincaid readability score \\
& n\_v & Noun-verb sequences \\
& n\_aux & Noun-auxiliary sequences \\
& n\_3s\_v & Third singular noun-verb sequences \\
& det\_n\_pl & Determiner-noun-pronoun sequences \\
& det\_pl\_n & Determiner-pronoun-noun sequences \\
& pro\_aux & Pronoun-auxiliary sequences \\
& pro\_3s\_v & Third singular pronoun-verb sequences \\
& total\_error & Total morphosyntactic errors \\
& total\_syl & Total syllable count \\
& average\_syl & Average syllables per word \\
& mlu\_words & Mean Length of Utterance in words \\
& mlu\_morphemes & Mean Length of Utterance in morphemes \\
& mlu100\_utts & MLU for first 100 utterances \\
& verb\_utt & Number of verb utterances \\
& dss & Developmental Sentence Score \\
\hline
\end{tabular}
\label{tab:features5}
\end{center}
\vspace{-4mm}
\end{table*}

\begin{table*}[!t]
\caption{Morphological Development Features}
\begin{center}
\begin{tabular}{|p{2.5cm}|p{4cm}|p{9cm}|}
\hline
\textbf{Category} & \textbf{Feature} & \textbf{Description} \\
\hline
\multirow{14}{*}{\parbox{2.5cm}{Morphological Development Markers}} 
& present\_progressive & Use of -ing forms \\
& propositions\_in & Usage of 'in' prepositions \\
& propositions\_on & Usage of 'on' prepositions \\
& plural\_s & Regular plural markers \\
& irregular\_past\_tense & Irregular past tense forms \\
& possessive\_s & Possessive markers \\
& uncontractible\_copula & Uncontracted forms of 'be' \\
& articles & Use of articles (a, an, the) \\
& regular\_past\_ed & Regular past tense markers \\
& regular\_3rd\_person\_s & Regular third person singular forms \\
& irregular\_3rd\_person & Irregular third person forms \\
& uncontractible\_aux & Uncontracted auxiliaries \\
& contractible\_copula & Contracted forms of 'be' \\
& contractible\_aux & Contracted auxiliaries \\
\hline
\end{tabular}
\label{tab:features4}
\end{center}
\vspace{-4mm}
\end{table*}
\begin{table*}[!t]
\caption{Language Model and Utterance Structure Features}
\begin{center}
\begin{tabular}{|p{2.5cm}|p{2cm}|p{11cm}|}
\hline
\textbf{Category} & \textbf{Feature} & \textbf{Description} \\
\hline
\multirow{4}{*}{\parbox{2.5cm}{Utterance Structure and Fluency}} 
& n\_dos & Number of 'do' auxiliaries used \\
& repetition & Count of word/phrase repetitions \\
& retracing & Number of retraced (self-corrected) utterances \\
& fillers & Count of filler words (um, uh, etc.) \\
\hline
\multirow{6}{*}{\parbox{2.5cm}{Language Model Perplexity}} 
& s\_1g\_ppl & 1-gram perplexity compared to SLI language model \\
& s\_2g\_ppl & 2-gram perplexity compared to SLI language model \\
& s\_3g\_ppl & 3-gram perplexity compared to SLI language model \\
& d\_1g\_ppl & 1-gram perplexity compared to TD language model \\
& d\_2g\_ppl & 2-gram perplexity compared to TD language model \\
& d\_3g\_ppl & 3-gram perplexity compared to TD language model \\
\hline
\end{tabular}
\label{tab:features2}
\end{center}
\vspace{-4mm}
\end{table*}

\begin{table*}[!t]
\caption{Z-Score Comparison Features}
\begin{center}
\begin{tabular}{|p{2.5cm}|p{2cm}|p{11cm}|}
\hline
\textbf{Category} & \textbf{Feature} & \textbf{Description} \\
\hline
\multirow{8}{*}{\parbox{2.5cm}{Z-Score Comparisons}} 
& z\_mlu\_sli & Z-score of Mean Length of Utterance relative to SLI group \\
& z\_mlu\_td & Z-score of MLU relative to TD group \\
& z\_word\_errors\_sli & Z-score of word errors relative to SLI group \\
& z\_word\_errors\_td & Z-score of word errors relative to TD group \\
& z\_r\_2\_i\_verbs\_sli & Z-score of verb ratio relative to SLI group \\
& z\_r\_2\_i\_verbs\_td & Z-score of verb ratio relative to TD group \\
& z\_utts\_sli & Z-score of utterances relative to SLI group \\
& z\_utts\_td & Z-score of utterances relative to TD group \\
\hline
\end{tabular}
\label{tab:features3}
\end{center}
\vspace{-4mm}
\end{table*}

\begin{table*}[!t]
\caption{Basic Production and Lexical Features}
\begin{center}
\begin{tabular}{|p{2.5cm}|p{2cm}|p{11cm}|}
\hline
\textbf{Category} & \textbf{Feature} & \textbf{Description} \\
\hline
\multirow{3}{*}{\parbox{2.5cm}{Basic Production Measures}} 
& child\_TNW & Total Number of Words produced by the child \\
& child\_TNS & Total Number of Sentences in the child's narrative \\
& examiner\_TNW & Total Number of Words spoken by the examiner, indicating level of support needed \\
\hline
\multirow{4}{*}{\parbox{2.5cm}{Lexical Diversity and Complexity}} 
& freq\_ttr & Frequency of Word Types to Word Token Ratio, measuring lexical diversity \\
& r\_2\_i\_verbs & Ratio of raw to inflected verbs, indicating morphological complexity \\
& mor\_words & Number of words in the morphological tier \\
& num\_pos\_tags & Number of different Part-of-Speech tags used \\
\hline
\end{tabular}
\label{tab:features1}
\end{center}
\vspace{-4mm}
\end{table*}

\begin{figure}[htbp]
    \centering
    \includegraphics[width=1\linewidth]{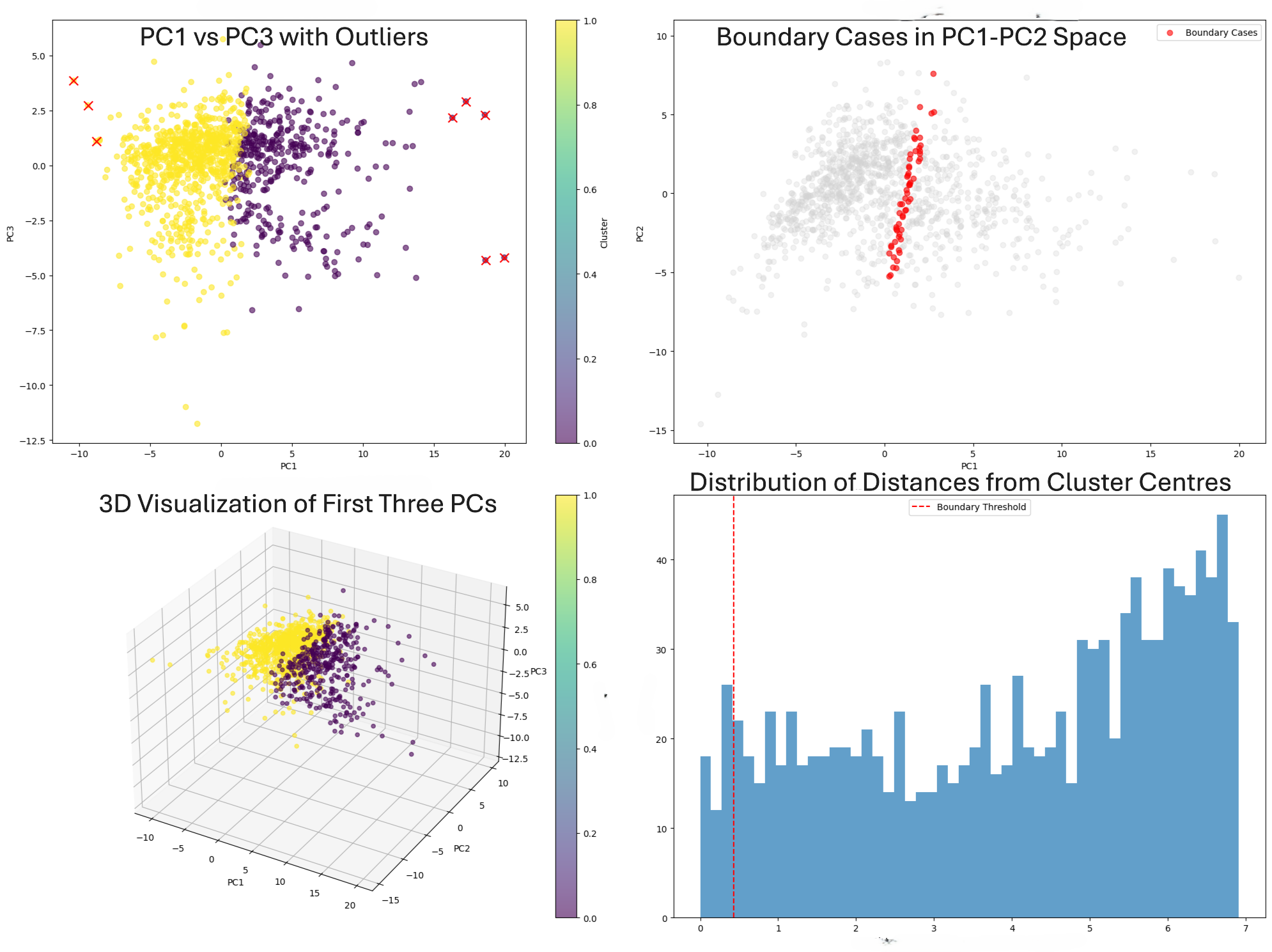}
    \caption{Detailed Analysis of Cluster Structure and Boundaries. This figure presents four complementary visualizations of the clustering results. (Top Left) PC1 vs PC2 plot with outliers marked as red X's shows the cluster separation with identified statistical outliers. (Top Right) Boundary cases (red points) plotted in PC1-PC2 space against the full dataset (gray points) demonstrate the transition zone between clusters. (Bottom Left) 3D visualization of the first three principal components reveals the complete spatial distribution of the clusters (purple and yellow), showing how the groups separate in three-dimensional space. (Bottom Right) Histogram of distances from cluster centers, with the boundary threshold (red dashed line) marking the 5th percentile of distance differences, used to identify boundary cases. The distribution shows a right-skewed pattern, indicating that most points have clear cluster membership while a smaller subset lies in the boundary region.}

    \label{fig:enter-label}
\end{figure}

\subsection{Cluster Analysis Results}

Silhouette analysis identified an optimal two-cluster solution with a maximum score of 0.36 and a secondary peak at k=5 (score $\approx 0.33$). The two clusters showed these characteristics:

\begin{itemize}
    \item \textbf{Cluster 0:} 373 participants (32.1\%), high positive PC1 values (mean = 4.61), negative PC2 values (mean = -0.88), and 17\% SLI prevalence. This cluster was characterized by high positive values on PC1 (mean = 4.61) and negative values on PC2 (mean = -0.88). This cluster demonstrated higher overall language production but lower syntactic complexity. Only 17\% of participants in this cluster were identified as having specific language impairment (SLI), suggesting a profile of robust verbal output with reduced complexity.
    \item \textbf{Cluster 1:} 790 participants (67.9\%), negative PC1 values (mean = -2.18), positive PC2 values (mean = 0.41), and 26\% SLI prevalence. This larger cluster, containing 790 participants (67.9\% of the sample), exhibited negative values on PC1 (mean = -2.18) and positive values on PC2 (mean = 0.41), indicating moderate language production coupled with higher syntactic complexity. This cluster had a higher prevalence of SLI, with 26\% of participants classified as having language impairment. These cluster characteristics are summarized in Table XIII.
\end{itemize}

\subsection{Boundary Case Identification}
The analysis identified 59 boundary cases (5.1\% of the sample) with intermediate characteristics between the main clusters. These cases showed lower SLI prevalence (11.9\%) compared to both main clusters, with mean PC1 value of 1.244 ± 0.606.

\begin{table}[!htbp]
\caption{Top Contributing Features for First Five Principal Components}
\centering
\small
\begin{tabular}{|l|l|l|}
\hline
\textbf{Component} & \textbf{Feature} & \textbf{Loading} \\
\hline
\multirow{5}{*}{PC1} & mor\_words & 0.240 \\
& child\_TNW & 0.239 \\
& total\_syl & 0.238 \\
& child\_TNS & 0.225 \\
& z\_utts\_td & 0.225 \\
\hline
\multirow{5}{*}{PC2} & mlu\_morphemes & 0.319 \\
& z\_mlu\_td & 0.315 \\
& z\_mlu\_sli & 0.315 \\
& mlu\_words & 0.315 \\
& verb\_utt & 0.293 \\
\hline
\multirow{5}{*}{PC3} & z\_word\_errors\_sli & 0.284 \\
& z\_word\_errors\_td & 0.284 \\
& word\_errors & 0.284 \\
& d\_2g\_ppl & 0.270 \\
& d\_3g\_ppl & 0.268 \\
\hline
\multirow{5}{*}{PC4} & z\_word\_errors\_td & 0.337 \\
& word\_errors & 0.337 \\
& z\_word\_errors\_sli & 0.337 \\
& regular\_3rd\_person\_s & 0.244 \\
& n\_3s\_v & 0.228 \\
\hline
\multirow{5}{*}{PC5} & r\_2\_i\_verbs & 0.286 \\
& z\_r\_2\_i\_verbs\_td & 0.286 \\
& z\_r\_2\_i\_verbs\_sli & 0.286 \\
& regular\_3rd\_person\_s & 0.244 \\
& n\_3s\_v & 0.232 \\
\hline
\multicolumn{3}{|l|}{\textsuperscript{a}TNW: Total Number of Words} \\
\multicolumn{3}{|l|}{\textsuperscript{b}TNS: Total Number of Syllables} \\
\multicolumn{3}{|l|}{\textsuperscript{c}SLI: Specific Language Impairment, TD: Typically Developing} \\
\hline
\end{tabular}
\label{tab:loadings}
\end{table}

\begin{table}[!htbp]
\renewcommand{\arraystretch}{1.3}
\caption{Descriptive Statistics of Principal Components (PC1-PC5)}
\label{tab:pc_stats}
\centering
\begin{tabular}{|l|r|r|r|r|r|}
\hline
\textbf{Statistic} & \textbf{PC1} & \textbf{PC2} & \textbf{PC3} & \textbf{PC4} & \textbf{PC5} \\
\hline
Mean & -9.78e-17 & 2.93e-16 & 1.47e-16 & -6.11e-17 & 4.15e-16 \\
\hline
SD & 4.09 & 2.80 & 2.01 & 1.82 & 1.73 \\
\hline
Min & -10.39 & -14.60 & -11.76 & -11.65 & -3.95 \\
\hline
25\% & -2.77 & -1.75 & -0.80 & -0.70 & -0.96 \\
\hline
Median & -0.56 & -0.01 & 0.47 & 0.17 & -0.20 \\
\hline
75\% & 2.03 & 1.93 & 1.21 & 0.86 & 0.80 \\
\hline
Max & 19.97 & 9.77 & 5.74 & 5.84 & 16.81 \\
\hline
\end{tabular}
\footnotetext{Components exhibit standardized characteristics with near-zero means and decreasing standard deviations (PC1: 4.09 to PC5: 1.73). PC1 (-10.39 to 19.97) and PC2 (-14.60 to 9.77) show the widest value ranges, capturing primary data variability. Symmetric distributions around zero, as shown by median and quartile values, confirm successful PCA standardization.}
\end{table}

\begin{table}[!htbp]
\caption{Principal Component Analysis Results}
\centering
\begin{tabular}{|c|c|r|r|r|}
\hline
\textbf{} & \textbf{Component} & \textbf{Eigenvalue} & \textbf{Variance (\%)} & \textbf{Cumulative (\%)} \\
\hline
0 & PC1\textsuperscript{a} & 3.97 & 28.35 & 28.35 \\
1 & PC2\textsuperscript{a} & 1.85 & 13.23 & 41.58 \\
2 & PC3\textsuperscript{b} & 0.96 & 6.87 & 48.46 \\
3 & PC4\textsuperscript{b} & 0.79 & 5.64 & 54.09 \\
4 & PC5\textsuperscript{b} & 0.71 & 5.05 & 59.14 \\
5 & PC6\textsuperscript{b} & 0.62 & 4.46 & 63.61 \\
6 & PC7\textsuperscript{b} & 0.57 & 4.05 & 67.66 \\
7 & PC8\textsuperscript{b} & 0.48 & 3.42 & 71.08 \\
8 & PC9\textsuperscript{b} & 0.37 & 2.63 & 73.71 \\
9 & PC10\textsuperscript{b} & 0.32 & 2.30 & 76.01 \\
10 & PC11\textsuperscript{b} & 0.30 & 2.16 & 78.17 \\
11 & PC12\textsuperscript{b} & 0.27 & 1.93 & 80.10 \\
12 & PC13\textsuperscript{b} & 0.25 & 1.75 & 81.85 \\
13 & PC14\textsuperscript{b} & 0.24 & 1.70 & 83.55 \\
\hline
\multicolumn{5}{|l|}{\textsuperscript{a}Components with eigenvalues $>$ 1 (retained)} \\
\multicolumn{5}{|l|}{\textsuperscript{b}Components with eigenvalues $<$ 1 (not retained)} \\
\hline
\end{tabular}
\label{tab:pca_results}
\end{table}

\begin{table}[!htbp]
\renewcommand{\arraystretch}{1.3}
\caption{Statistical Summary of Boundary Cases (n = 59)}
\label{tab:boundary_stats}
\centering
\footnotesize
\begin{tabular}{|l|c|p{3.5cm}|}
\hline
\textbf{Comp.} & \textbf{Mean $\pm$ SD} & \textbf{Clinical Comparison} \\
\hline
PC1 & 1.24 $\pm$ 0.61 & Between Cl.0 (4.61) and Cl.1 (-2.18) \\
\hline
PC2 & -0.09 $\pm$ 3.01 & Near-neutral position \\
\hline
PC3 & 0.24 $\pm$ 1.89 & Similar to cluster means \\
\hline
SLI & 11.9\% & Lower than both clusters \\
\hline
\end{tabular}
\end{table}

\begin{table}[!htbp]
\renewcommand{\arraystretch}{1.3}
\caption{Summary Statistics of Cluster Characteristics}
\label{tab:cluster_stats}
\centering
\begin{tabular}{|c|c|c|c|c|c|}
\hline
\textbf{Cluster} & \textbf{Size} & \textbf{PC1\_mean} & \textbf{PC2\_mean} & \textbf{PC3\_mean} & \textbf{Y\_ratio} \\
\hline
0 & 373 & 4.610 & -0.875 & -0.061 & 0.166 \\
\hline
1 & 790 & -2.177 & 0.413 & 0.029 & 0.259 \\
\hline
\end{tabular}
\footnotetext{For each cluster (0 and 1), the table shows the cluster size, mean values of the first three principal components (PC1-PC3), and the proportion of cases with language impairment (Y\_ratio). Cluster 0 (n=373) is characterized by high positive PC1 values and lower impairment rates (16.6\%), while Cluster 1 (n=790) shows negative PC1 values and higher impairment rates (25.9\%). The contrasting PC1 means ($+$4.610 vs $-$2.177) indicate that language production is the primary differentiator between clusters.}
\end{table}

\begin{table}[!htbp]
\renewcommand{\arraystretch}{1.3}
\caption{Descriptive Statistics of Language Features}
\label{tab:descriptive_stats}
\centering
\begin{tabular}{|l|r|r|r|r|r|}
\hline
\textbf{Feature} & \textbf{Mean} & \textbf{SD} & \textbf{Min} & \textbf{Max} & \textbf{N} \\
\hline
child\_TNW & 423.15 & 242.68 & 29.00 & 1746.00 & 1163.0 \\
\hline
child\_TNS & 50.41 & 27.54 & 7.00 & 189.00 & 1163.0 \\
\hline
word\_errors & 0.56 & 1.45 & 0.00 & 16.00 & 1163.0 \\
\hline
mlu\_morphemes & 8.49 & 1.85 & 2.70 & 18.00 & 1163.0 \\
\hline
total\_syl & 497.00 & 288.36 & 30.00 & 2119.00 & 1163.0 \\
\hline
examiner\_TNW & 13.38 & 30.69 & 0.00 & 621.00 & 1163.0 \\
\hline
freq\_ttr & 0.37 & 0.11 & 0.15 & 0.83 & 1163.0 \\
\hline
\end{tabular}
\footnotetext{TNW = Total Number of Words; TNS = Total Number of Sentences; MLU = Mean Length of Utterance; TTR = Type-Token Ratio.}
\end{table}

\begin{table}[!htbp]
\renewcommand{\arraystretch}{1.3}
\caption{Feature Loadings for First Three Principal Components (Top 20 Features by PC1 Magnitude)}
\label{tab:feature_loadings_top20}
\centering
\begin{tabular}{|l|r|r|r|}
\hline
\textbf{Feature} & \textbf{PC1} & \textbf{PC2} & \textbf{PC3} \\
\hline
mor\_words & \textbf{0.240} & -0.001 & 0.008 \\
\hline
child\_TNW & \textbf{0.239} & -0.011 & 0.021 \\
\hline
total\_syl & \textbf{0.238} & 0.007 & 0.038 \\
\hline
child\_TNS & \textbf{0.225} & -0.111 & 0.039 \\
\hline
z\_utts\_td & \textbf{0.225} & -0.111 & 0.039 \\
\hline
z\_utts\_sli & \textbf{0.225} & -0.111 & 0.039 \\
\hline
mlu100\_utts & \textbf{0.222} & -0.105 & 0.026 \\
\hline
ipsyn\_total & \textbf{0.207} & 0.070 & 0.087 \\
\hline
total\_error & \textbf{0.206} & -0.029 & -0.051 \\
\hline
freq\_ttr & \textbf{-0.202} & 0.045 & 0.070 \\
\hline
uncontractible\_copula & 0.198 & -0.005 & 0.074 \\
\hline
present\_progressive & 0.182 & -0.077 & -0.130 \\
\hline
uncontractible\_aux & 0.178 & -0.076 & -0.113 \\
\hline
articles & 0.173 & -0.133 & -0.088 \\
\hline
num\_pos\_tags & 0.164 & 0.093 & 0.152 \\
\hline
retracing & 0.160 & -0.117 & -0.108 \\
\hline
propositions\_on & 0.158 & 0.011 & -0.049 \\
\hline
irregular\_past\_tense & 0.154 & 0.077 & 0.196 \\
\hline
pro\_3s\_v & 0.150 & -0.014 & 0.117 \\
\hline
det\_n\_pl & 0.145 & 0.098 & 0.062 \\
\hline
\end{tabular}
\footnotetext{Bold values indicate loadings > |0.2|. Features are sorted by absolute magnitude of PC1 loadings. PC = Principal Component. Variable prefixes: z\_ = standardized score, mlu = mean length of utterance, TNW = total number of words.}
\end{table}

\begin{table}[!htbp]
\renewcommand{\arraystretch}{1.15}
\caption{Clinical Features by Cluster}
\label{tab:clinical_features}
\centering
\begin{tabular}{|l|c|c|c|c|}
\hline
\textbf{Feature} & \textbf{Cluster 0} & \textbf{Cluster 1} & \textbf{\textit{p}-value} & \textbf{Cohen's \textit{d}} \\
\hline
child\_TNW & 690.539 & 296.897 & 0.000 & 2.483 \\
\hline
mlu\_morphemes & 8.586 & 8.448 & 0.238 & 0.074 \\
\hline
word\_errors & 0.260 & 0.704 & 0.000 & -0.310 \\
\hline
\multicolumn{5}{|l|}{\textsuperscript{a}TNW = Total Number of Words; MLU = Mean Length of Utterance} \\
\hline
\end{tabular}
\end{table}

\begin{table}[!htbp]
\renewcommand{\arraystretch}{1.15}
\caption{Cluster Agreement Metrics Across Different Principal Component Spaces}
\label{tab:cluster_agreement}
\centering
\begin{tabular}{|l|c|c|c|}
\hline
\textbf{Metric} & \textbf{PC1-PC2} & \textbf{PC1-PC2} & \textbf{PC1-PC3} \\
& \textbf{vs PC1-PC3} & \textbf{vs PC2-PC3} & \textbf{vs PC2-PC3} \\
\hline
Adjusted Rand & 0.861 & 0.054 & 0.026 \\
Index & & & \\
\hline
Adjusted Mutual & 0.758 & 0.039 & 0.017 \\
Information & & & \\
\hline
Accuracy & 0.965 & 0.617 & 0.582 \\
(best mapping) & & & \\
\hline
\end{tabular}
\end{table}

\section{Analysis and Discussion}

\subsection{Interpretation of Principal Component Structure}
The three principal components reveal a hierarchical organization of language abilities. PC1 (28.35\% of variance) primarily represents overall language production capacity, with its strong loadings on morphological words, total word count, and syllables. This suggests that quantitative output forms the primary axis of variation in language development.

PC2 (13.23\% of variance) captures syntactic complexity independent of production volume, representing a qualitative dimension of language ability through its loadings on MLU measures and verb usage patterns. PC3 (6.87\% of variance) reveals an accuracy dimension through error patterns and perplexity scores, operating independently of both volume and complexity.

This three-dimensional structure suggests language development follows distinct but related pathways of quantity (production), quality (complexity), and accuracy (error patterns).

\subsection{Cluster Characteristics and Clinical Significance}
The two-cluster solution reveals distinct linguistic profiles with significant clinical implications:

Cluster 0's high production but lower SLI prevalence (16.6\%) contrasts with Cluster 1's reduced production and higher SLI prevalence (25.9\%). Statistical comparisons between clusters showed significant differences in total word production ($p < 0.001$, Cohen's $d = 2.483$) and error rates ($p < 0.001$, Cohen's $d = -0.310$), but no significant difference in MLU morphemes ($p = 0.238$, Cohen's $d = 0.074$).

These findings challenge traditional SLI assessment approaches that emphasize syntactic complexity (measured through MLU). Instead, these results suggest SLI manifests primarily as reduced production capacity, with error patterns as secondary indicators.

\subsection{Significance of Boundary Cases}
The boundary cases with intermediate characteristics and lower SLI prevalence (11.9\%) strongly support a continuous model of language ability rather than discrete categories. The transitional zone in language production capacity (intermediate PC1 values) alongside high PC2 variance (±3.010) indicates that syntactic complexity varies independently of boundary status.

This finding has profound implications for clinical assessment, suggesting diagnostic approaches should avoid rigid categorical distinctions in favor of multidimensional profiles that acknowledge the continuous nature of language abilities.

\subsection{Clinical Applications}
The analysis indicates several directions for improving clinical assessment of SLI:

\begin{itemize}
    \item Greater emphasis on production volume and error rates in diagnostic protocols
    \item Development of multidimensional assessment tools that capture the three primary dimensions identified (production, complexity, and accuracy)
    \item Consideration of gradient or spectrum-based diagnostic frameworks rather than binary classifications
    \item Targeted interventions based on specific deficit patterns across the three dimensions
\end{itemize}

The strong stability of PC1-based clustering emphasizes the importance of production measures in clinical evaluation, though the presence of boundary cases cautions against rigid categorical distinctions in assessment protocols.

\section{Limitations and Future Work}

This study has several key limitations. First, although PCA is effective for dimensionality reduction, the first three principal components explained only 48.46\% of the total variance, indicating substantial unexplained variability. This is typical for complex linguistic datasets but remains a limitation of the approach. Second, the 5th percentile threshold chosen for identifying boundary cases, while methodologically sound, was a somewhat arbitrary cutoff that could yield different results if altered.

The dataset, though relatively large and diverse, is cross-sectional, limiting the ability to draw conclusions about developmental trajectories. The age distribution provides useful insights, but true developmental patterns require longitudinal tracking. Additionally, differences in elicitation protocols across the three corpora may introduce methodological variability despite standardization efforts.

Future research should pursue longitudinal studies to track how these linguistic profiles evolve over time, especially for children in the boundary cases. Expanding feature sets to include pragmatic and discourse-level measures could capture additional dimensions of language impairment not represented in the current analysis. Cross-linguistic applications would help identify universal versus language-specific patterns in SLI manifestation.

Further research is also needed to validate the clinical utility and actionability of these findings. This should include investigating how these profiles relate to other clinical measures, how they predict treatment outcomes, and whether targeted interventions based on cluster membership yield improved results. The identified multidimensional structure could inform more nuanced assessment tools that move beyond categorical diagnoses toward profile-based approaches.

\section{Conclusion}  
These findings suggest that language impairment is best understood as a multidimensional construct rather than a uniform deficit pattern. The identification of two distinct clusters with different linguistic profiles and varying SLI prevalence rates underscores the complexity of language impairment. Furthermore, the presence of boundary cases highlights a continuum of language abilities, emphasizing the need for nuanced diagnostic and intervention approaches.

\end{document}